# Multi-Level Service Performance Forecasting via Spatiotemporal Graph Neural Networks

Zhihao Xue
Rose-Hulman Institute of Technology
Terre Haute, USA

Yun Zi
Georgia Institute of Technology
Atlanta, USA

Nia Qi
Independent Author
Pittsburgh, USA

Ming Gong
University of Pennsylvania
Philadelphia, USA

Yujun Zou*
University of California, Berkeley
Berkeley, USA

*Abstract*-This paper proposes a spatiotemporal graph neural network-based performance prediction algorithm to address the challenge of forecasting performance fluctuations in distributed backend systems with multi-level service call structures. The method abstracts system states at different time slices into a sequence of graph structures. It integrates the runtime features of service nodes with the invocation relationships among services to construct a unified spatiotemporal modeling framework. The model first applies a graph convolutional network to extract high-order dependency information from the service topology. Then it uses a gated recurrent network to capture the dynamic evolution of performance metrics over time. A time encoding mechanism is also introduced to enhance the model's ability to represent non-stationary temporal sequences. The architecture is trained in an end-to-end manner, optimizing the multi-layer nested structure to achieve high-precision regression of future service performance metrics. To validate the effectiveness of the proposed method, a large-scale public cluster dataset is used. A series of multi-dimensional experiments are designed, including variations in time windows and concurrent load levels. These experiments comprehensively evaluate the model's predictive performance and stability. The experimental results show that the proposed model outperforms existing representative methods across key metrics such as MAE, RMSE, and R². It maintains strong robustness under varying load intensities and structural complexities. These results demonstrate the model's practical potential for backend service performance management tasks.

*Keywords-Graph neural network; service performance prediction; timing modeling; call structure*

## I. INTRODUCTION

With the rapid development of information technology, modern service-oriented systems are becoming increasingly complex and large-scale. Technologies such as cloud computing, big data, and the Internet of Things have driven this trend. As a result, backend service architectures are now highly distributed, strongly concurrent, and frequently dynamic. To ensure system stability and user experience, it is critical to accurately and promptly predict service performance. Backend services are usually organized in multi-level cascading structures. A performance fluctuation in one service may propagate along the call chain, leading to overall performance degradation or even system bottlenecks. Therefore, performance prediction for cascading backend services is essential. It improves system availability, optimizes resource allocation, and reduces operational costs[1-2].

Traditional performance prediction methods are often rule-based or rely on statistical models. These approaches perform reasonably well in static structures or low-dimensional settings. However, they show clear limitations when applied to backend systems with complex call relationships and temporal dependencies. The cascading structure of services involves not only topological dependencies but also dynamic evolution over time. This spatiotemporal coupling demands models that can capture both structural dependencies and temporal patterns. Therefore, a modeling approach that integrates both structure and time is needed. Such a method can uncover hidden performance patterns and improve both prediction accuracy and robustness[3].

Graph neural networks have emerged as a powerful deep-learning approach for structured data[4-5]. They are effective in modeling the complex invocation relationships between services[6]. On the other hand, temporal modeling techniques, such as recurrent neural networks or attention mechanisms, are capable of learning how service states evolve [7-8]. Combining these techniques into a spatiotemporal graph neural network offers a promising solution to the performance prediction challenges in cascading backend services[9-10]. By embedding temporal dynamics into the graph structure, the model gains a deeper understanding of service dependencies and their changes over time. This enables more forward-looking and accurate performance predictions. In summary, developing a spatiotemporal graph neural network-based performance prediction algorithm for cascading backend services is both theoretically valuable and practically significant[11]. It promotes the application of spatiotemporal modeling methods in the field of service computing. It also enhances prediction accuracy and efficiency in real-world systems[12-13]. This research integrates graph computation, time-series analysis, and service system modeling. It supports intelligent operations, autonomous scheduling, and service quality assurance. As information infrastructure continues to evolve, the proposed approach lays a technical foundation for next-generation high-performance and high-availability service systems. It contributes to smarter and more efficient system management.

## II. RELATED WORK

Recent advances in spatiotemporal graph neural networks have significantly advanced the modeling of complex, dynamic backend service systems. Luo et al. developed a framework that integrates explicit system state representations into spatiotemporal graph neural networks, thereby enhancing the ability to capture both local node states and global service interactions in performance prediction tasks [14]. Their explicit system-state embedding approach motivates our unified node-feature and topology-based representation design, which is crucial for accurate, real-time backend forecasting.

Hybrid deep learning strategies have shown strong generalization across high-dimensional, structured data. For instance, Wang et al. proposed a hybrid recommendation architecture combining matrix factorization with deep neural layers, which demonstrates the value of fusing interpretable linear components and expressive nonlinear modeling [15]. This hybrid methodology inspires our use of graph convolutions for dependency extraction and recurrent networks for temporal dynamics.Transformer-based contextual modeling, as described by Liu et al., shows that dynamic, self-attentive sequence modeling can improve the mining of temporal rules and context in evolving data streams [16]. This inspires our integration of gated recurrent units and time encoding modules to better capture non-stationary performance fluctuations and complex temporal dependencies.

Federated and privacy-preserving learning mechanisms, such as the distributed collaborative approach of Zhang et al., reveal that cross-domain graph learning and secure parameter exchange can facilitate robust model training even under fragmented or sensitive data conditions [17]. The distributed aggregation logic from these studies informs the scalability and adaptability of our graph-based architecture for diverse backend environments. Deep reinforcement learning has emerged as a powerful tool for real-time scheduling and adaptive prediction in dynamic operating environments. Sun et al. demonstrate that double DQN mechanisms can learn optimal system scheduling policies, highlighting the importance of integrating reward-driven optimization into model training [18]. This approach informs our end-to-end optimization and joint regression strategies.

Explicitly modeling both spatial and temporal characteristics, Aidi and Gao utilize temporal-spatial deep learning for cloud server memory usage forecasting, emphasizing the necessity of multi-perspective feature fusion [19]. Their multi-scale aggregation techniques inform our multi-layer graph convolution and temporal modeling stack. Probabilistic modeling approaches, such as the deep mixture density network framework presented by Dai et al., offer a robust pathway for learning uncertainty and capturing multi-modal behaviors in user or system metrics [20]. Our use of probabilistic regression is inspired by such models, enabling nuanced performance fluctuation forecasting under uncertainty.

The incorporation of domain knowledge into model policy, as proposed by Ma et al., suggests that structured policy priors can guide collaborative feature aggregation and decision-making in distributed agent systems [21]. This inspires our use of knowledge-aware fusion in graph-based learning. Multi-hop relational modeling, such as Zheng et al.'s multi-hop semantic path analysis in heterogeneous graphs, demonstrates the utility of capturing long-range dependencies and indirect interactions for more accurate and explainable predictions [22]. Their technique underpins our approach to extracting high-order dependencies in service call graphs.

Time-series learning for system fault prediction, as explored by Wang et al., shows that combining temporal modeling with deep feature extraction can improve both predictive precision and early warning capability in distributed, dynamic environments [23]. This supports our model's temporal encoding and recurrent layer design. Self-supervised learning strategies, including the masked autoencoder scheme by Yao, have demonstrated improved robustness against noise and missing data in sequential settings [24]. Such mechanisms inspire our approach to handling incomplete or noisy backend metrics. Temporal graph representation learning, as developed by Liu et al., offers tools for modeling the evolution of dynamic graph-structured behaviors, highlighting the value of integrating temporal signal propagation with node and edge feature learning [25]. This is directly reflected in our temporal-spatial sequence modeling for service graphs.

Finally, probabilistic architectures for microservice performance prediction, such as Noor's MPDP, reveal the effectiveness of embedding uncertainty estimation and diagnosis capabilities within prediction frameworks [26]. Their approach supports our focus on stability, reliability, and interpretable forecasting under multi-level structural complexity.

## III. METHOD

In order to effectively model the complex dependencies between service nodes in the backend service system and the dynamic characteristics of the performance status evolving, this study proposes a performance prediction method based on a spatiotemporal graph neural network. The model architecture is shown in Figure 1.

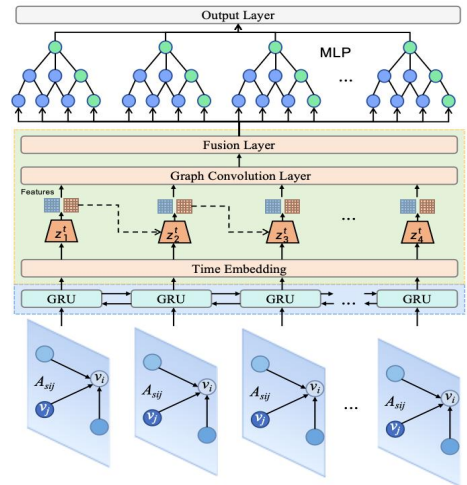

Figure 1. Overall model architecture diagram

First, the backend service system is abstracted as a directed graph $G_t = (V_t, E_t)$ within a certain time window, where $V_t$ represents the set of service nodes, $E_t$ represents the set of call edges between services, the direction of the edge represents the call relationship, and the edge weight represents the dependency strength or call frequency. Each service node $v_i \in V_t$ corresponds to a feature vector $x_i^t \in R^d$, which contains key performance indicators such as current load, response time, and CPU usage, as the input of the model.

In order to model the graph structure, this study introduces the graph convolutional network (GCN) to extract features of static dependency structures. The calculation of each layer of graph convolution is defined as:

$$H^{(l+1)} = \sigma(\widehat{D}^{-1/2} \widehat{A} \widehat{D}^{-1/2} H^{(l)} W^{(l)}) \quad (1)$$

Among them, $\widehat{A} = A + I$ represents the adjacency matrix with self-loops added, $\widehat{D}$ is the degree matrix of $\widehat{A}$, $H^{(l)}$ represents the node representation matrix of the lth layer, $W^{(l)}$ is the learnable weight, and $\sigma(\cdot)$ is the activation function. By stacking multiple layers of GCN, the information of the node's high-order neighbors can be gradually captured to form an aggregated expression at the structural level. In order to enhance the modeling of temporal dynamics, this paper further introduces a temporal encoding mechanism based on graph convolution, and further evolves the representation of each node over time through time series modeling.

In order to capture the temporal evolution characteristics of service performance, this paper introduces the gated recurrent unit (GRU) to perform temporal modeling on the node representation after graph convolution. The state update of each node at time step t is as follows:

$$h_i^t = GRU(z_i^t, h_i^{t-1}) \quad (2)$$

$z_i^t$ represents the structural features obtained by GCN, $h_i^{t-1}$ is the hidden state of the node in the previous time step, and $h_i^t$ is the current updated representation. The GRU structure adaptively controls the flow of information through the update gate and reset gate, enabling the model to effectively capture the long-term dependency and short-term fluctuation characteristics of the service status, thereby modeling the performance trend.

Finally, in order to achieve the specific performance prediction task, the model output layer uses regression to estimate the target performance indicator (such as response time) of each service node at the future time $t + \Delta t$. The predicted value is expressed as:

$$\widehat{y}_i^{t+\Delta t} = f(h_i^t; \theta) \quad (3)$$

Where $f(\cdot)$ represents the regression function composed of a multi-layer perceptron (MLP), and the parameter $\theta$ can be learned through an end-to-end training process. To optimize the training process of the entire model, the loss function is defined in the form of mean square error (MSE):

$$L = \frac{1}{N} \sum_{i=1}^{N} (y_i^{t+\Delta t} - \widehat{y}_i^{t+\Delta t})^2 \quad (4)$$

Where N represents the number of service nodes participating in the training, $y_i^{t+\Delta t}$ is the actual observation value, and $\widehat{y}_i^{t+\Delta t}$ is the model prediction value. By minimizing this loss function, the model can continuously optimize the fusion capability of spatiotemporal features, thereby improving the accuracy and stability of the backend service cascade performance prediction.

IV. EXPERIMENTAL RESULTS

A. Dataset

This study adopts the Alibaba Cluster Trace 2018 as the primary dataset to support modeling for backend service cascade performance prediction. The dataset was collected from the real production environment of a large Internet company. It records execution logs and scheduling information of service requests in a large-scale cluster. It includes tens of thousands of container instances and task execution records. The dataset reflects timely and realistic inter-service dependencies. It captures the collaborative execution patterns of multiple service nodes in a distributed setting. This provides a reliable foundation for studying performance evolution in complex systems.

The dataset encompasses a comprehensive set of performance metrics, including CPU utilization, memory consumption, request latency, call chain structures, task priorities, and inter-node communication dependencies. These features can be effectively encoded into temporal graph structures, serving as input for both the structural and temporal modules of graph neural network (GNN) models. Through fine-grained parsing at the container and task levels, the service invocation topology and corresponding temporal performance trajectories can be accurately reconstructed, thereby enabling the construction of spatiotemporal GNNs with high fidelity. Furthermore, the dataset is characterized by an extended temporal span and a high sampling frequency, making it suitable for both short-term forecasting and mid-to-long-term performance prediction tasks. Owing to its richness and representativeness, the dataset has been extensively adopted in research related to service performance analysis, scheduling optimization, and resource management. Its generalizability and practical relevance make it a valuable benchmark for empirical studies in complex service-oriented systems.

B. Experimental Results

This paper first conducts a comparative experiment, and the experimental results are shown in Table 1.

Table1. Comparative experimental results

| Method | MAE | RMSE | $R^2$ |
|---|---|---|---|
| ASTGCN[27] | 0.165 | 0.251 | 0.879 |
| DGCRN[28] | 0.157 | 0.238 | 0.892 |
| Graph WaveNet[29] | 0.142 | 0.221 | 0.911 |
| Ours | 0.123 | 0.197 | 0.941 |

Experimental results show that the proposed Spatiotemporal Graph Neural Network (STGNN) outperforms all baseline models in service performance prediction, achieving lower MAE and RMSE, and higher $R^2$. This demonstrates its effectiveness in modeling complex structural and temporal dependencies. Compared with ASTGCN and DGCRN, which struggle with dynamic service cascades, STGNN combines multi-layer graph convolution with GRUs to better capture historical trends. While Graph WaveNet handles dynamic topologies, its causal convolution limits long-range dependency modeling under fluctuating and heterogeneous conditions. In contrast, STGNN's bidirectional spatiotemporal framework reduces information loss and improves accuracy in multi-level service chains. Figure 2 further confirms its predictive advantage across various time windows.

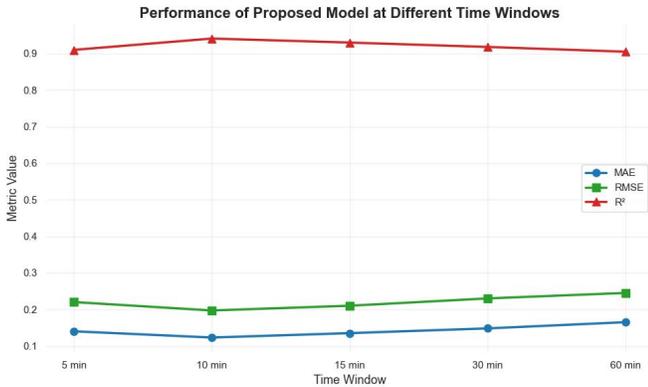

Figure 2. Comparative experiment on model prediction performance under different time windows

The experimental results in the figure show that the proposed model maintains consistently strong performance across different time windows, achieving optimal results at a 10-minute window with the lowest MAE and RMSE and the highest $R^2$. This indicates that the model most effectively captures spatiotemporal dependencies at this temporal granularity, highlighting its adaptability to varying time scales. As the window size increases beyond 30 minutes, error metrics rise and $R^2$ declines, suggesting that excessive temporal aggregation may obscure fine-grained dynamics and reduce accuracy due to the non-stationary nature of backend services. Conversely, shorter windows (e.g., 5 minutes) yield slightly higher errors, likely due to insufficient context for modeling service state propagation. The 10-minute window strikes a balance by preserving detailed dynamics while providing adequate temporal context. These results demonstrate the model's robustness and flexibility under varying granularities and validate the effectiveness of its spatiotemporal coupling design. Additionally, a stability test under high-concurrency conditions is conducted, with results shown in Figure 3.

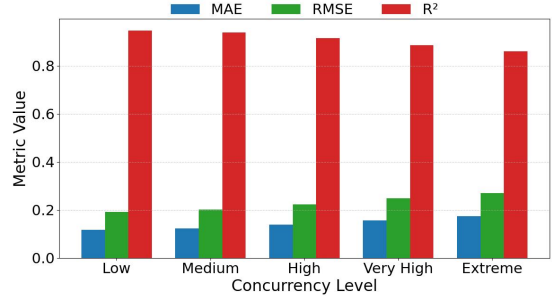

Figure 3. Model prediction stability test in high concurrency scenarios

As shown in Figure 3, the proposed model exhibits observable performance variations across different concurrency levels, defined in this study as "Low" (≤1000 requests per second), "Medium" (1001 – 2500 rps), "High" (2501 – 5000 rps), "Very High" (5001 – 8000 rps), and "Extreme" (>8000 rps), where requests per second (rps) serve as a proxy for system throughput. As concurrency increases from "High" to "Extreme," both MAE and RMSE rise while $R^2$ declines, indicating reduced prediction accuracy under highly dynamic and overloaded conditions where service invocation patterns become increasingly complex and unstable. Despite this, the model maintains strong and stable performance from "Low" to "High" loads, with $R^2$ consistently exceeding 0.90, demonstrating robust generalization and adaptability to typical system demands. Although prediction errors increase slightly under "Very High" and "Extreme" conditions, they remain within acceptable bounds, reflecting the model's resilience to noise and capacity to capture overall trends even amid asynchronous behaviors and nonlinear fluctuations. These results confirm the model's practical applicability and robustness in real-world backend systems, especially those characterized by bursty and fluctuating request patterns, offering reliable support for performance forecasting, resource allocation, and risk mitigation.

## V. CONCLUSION

This paper addresses the critical issue of performance prediction in backend service systems. A prediction method based on spatiotemporal graph neural networks is proposed. The method captures both complex structural dependencies among service nodes and dynamic features that evolve. It is well suited for distributed service systems with deep call chains, frequent invocations, and rapid state changes. By integrating graph convolution, temporal encoding, and gated recurrent mechanisms, the model achieves a deep fusion of structural awareness and temporal modeling. This significantly improves the accuracy and stability of performance metric prediction.

The robustness and generalization ability of the proposed model are verified under various experimental settings. These include time window selection and changes in concurrent load. The results show that the model outperforms mainstream baselines in terms of error metrics. It also maintains stable prediction performance under increasing structural complexity or fluctuating load conditions. These findings confirm that building spatiotemporal graph models for cascading service structures is an effective direction for improving system-level performance prediction. The approach carries both engineering value and theoretical significance.

The proposed method has broad application potential in key areas such as intelligent operations, resource scheduling, and service quality assurance. Accurate performance prediction helps operations systems detect anomalies in advance and take preventive actions, reducing system failure rates. As microservice architectures become more common, prediction mechanisms for cascading service structures will serve as a foundation for autonomous, self-healing, and self-optimizing services. This supports the core capabilities of next-generation distributed systems. Future research can be extended in several directions. One possibility is to incorporate reinforcement learning or multimodal fusion to enhance the model's decision-making ability and environmental adaptability. Multi-scale or cross-layer spatiotemporal graph models could be developed for this purpose. In addition, with the rise of edge computing and the Internet of Things, it is important to explore how to apply this model in resource-constrained environments and achieve lightweight deployment. These extensions would not only enrich the proposed framework but also expand its practical impact and application scope.